**AI-based framework to predict animal and pen feed intake in feedlot beef cattle**

Alex S. C. Maia[a*], John B. Hall[b], Hugo F. M. Milan[c], Izabelle A. M. A. Teixeira[d]

[a] Innovation and Sustainability in Animal Biometeorology (Inoβio), Laboratory of Animal Biometeorology, Department of Animal Science, State University of São Paulo, Jaboticabal Campus, 14844-900, SP, Brazil

[b] Department of Animal, Veterinary and Food Sciences, Nancy M. Cummings REEC, University of Idaho, Carmen, Idaho, USA

[c] Electrical Engineering Department, Federal University of Rondônia, Porto Velho/RO, Brazil

[d] Department of Animal, Veterinary and Food Sciences, Twin Falls Research and Extension Center, University of Idaho, 83301, Twins Fall, Idaho, USA

\* Corresponding author. E-mail address: alex.maia@unesp.br (A. S. C. Maia)

**Abstract:**

Advances in technology are transforming sustainable cattle farming practices, with electronic feeding systems generating big longitudinal datasets on individual animal feed intake, offering the possibility for autonomous precision livestock systems. However, the literature still lacks a methodology that fully leverages these longitudinal big data to accurately predict feed intake accounting for environmental conditions. To fill this gap, we developed an AI-based framework to accurately predict feed intake of individual animals and pen-level aggregation. Data from 19 experiments (>16.5M samples; 2013-2024) conducted at Nancy M. Cummings Research Extension & Education Center (Carmen, ID) feedlot facility and environmental data from AgriMet Network weather stations were used to develop two novel environmental indices: InComfort-Index, based solely on meteorological variables, showed good predictive capability for thermal comfort but had limited ability to predict feed intake; EASI-Index, a hybrid index integrating environmental variables with feed intake behavior, performed well in predicting feed intake but was less effective for thermal comfort. Together with the environmental indices, machine learning models were trained and the best-performing machine learning model (XGBoost) accuracy was RMSE of 1.38 kg/day for animal-level and only 0.14 kg/(day-animal) at pen-level. This approach provides a robust AI-based framework for predicting feed intake in individual animals and pens, with potential applications in precision management of feedlot cattle, through feed waste reduction, resource optimization, and climate-adaptive livestock management.

**Highlights**

- Novel AI-based framework to predict animal and pen-level feed intake

- Big data from electronic feeding systems with outlier removal using GAM and NMR

- Two new environmental indices: InComfort and EASI

- Previous days, weather, InComfort, and EASI as predictors of feed intake

- XGBoost RMSE of only 1.38 kg/day for animals and 0.14 kg/day-animal for pen-level

# 1. INTRODUCTION

Advancements in electronics, Big Data Analytics, Artificial Intelligence, Internet of Things, and wireless communication are transforming modern sustainable cattle farming practices (Aquilani et al., 2020; Fournel et al., 2017). In livestock operations, electronic feeding systems generate massive longitudinal datasets of individual animal feed intake (Simanungkalit et al., 2020), presenting both an opportunity and a challenge: how to leverage advanced AI-based methods to extract actionable insights from complex, high-dimensional temporal data while accounting for dynamic environmental conditions? Despite the availability of sophisticated gradient boosting algorithms and computational resources, a critical gap persists in developing robust AI-based prediction frameworks that accurately model feed intake as a function of cumulative environmental stressors—a problem with significant implications for precision livestock farming and sustainable food production (Wolfert et al., 2017).

Traditional thermal comfort indices, such as Temperature-Humidity Index (THI) and Comprehensions Climate Index (CCI; Hill and Wall, 2017; Mader et al., 2010), were developed in the pre-Big Data era, using limited datasets and linear statistical approaches. While these indices successfully capture short-term effects of thermal environmental conditions on physiological responses (Chang-Fung-Martel, et al., 2021), they fundamentally lack the ability to account for cumulative temporal effects.

In the current era of precision livestock farming, surprisingly, correlations between thermal environmental conditions and feed intake have been found weak (ArunKumar et al., 2025), which contradicts established understanding of thermal environmental effects on livestock (Collier et al., 2015, 2017; Gorczyca et al., 2018; Maia et al., 2023; Milan et al., 2019). These contradictions may result from not accounting for cumulative environmental conditions over preceding days or from not integrating environmental indices with physiological responses.

To address these computational and methodological gaps, we developed a novel AI-based framework to predict feed intake of feedlot cattle using 11 years of longitudinal data (>16.5M samples). Our novel contributions include 1) development of two conceptually distinct AI-based environmental indices (InComfort-Index, based solely on meteorological variables; EASI-Index, a hybrid index that integrates environmental variables with feed intake behavioral responses), 2) implementation of machine learning models to predict high-dimensional temporal feed intake data, and 3) validation of prediction accuracy. The proposed AI-based frameork provides high predictive performance and potential applications for real-time control systems in precision management of feedlot beef cattle, with direct application in feed waste reduction, resource optimization, and climate-adaptive livestock management.

## 2. MATERIALS AND METHODS

The AI-base framework is presented in Fig. 1. Data were processed using SAS and Python. All animal experiments were approved by the University of Idaho Institutional Animal Care and Use Committee.

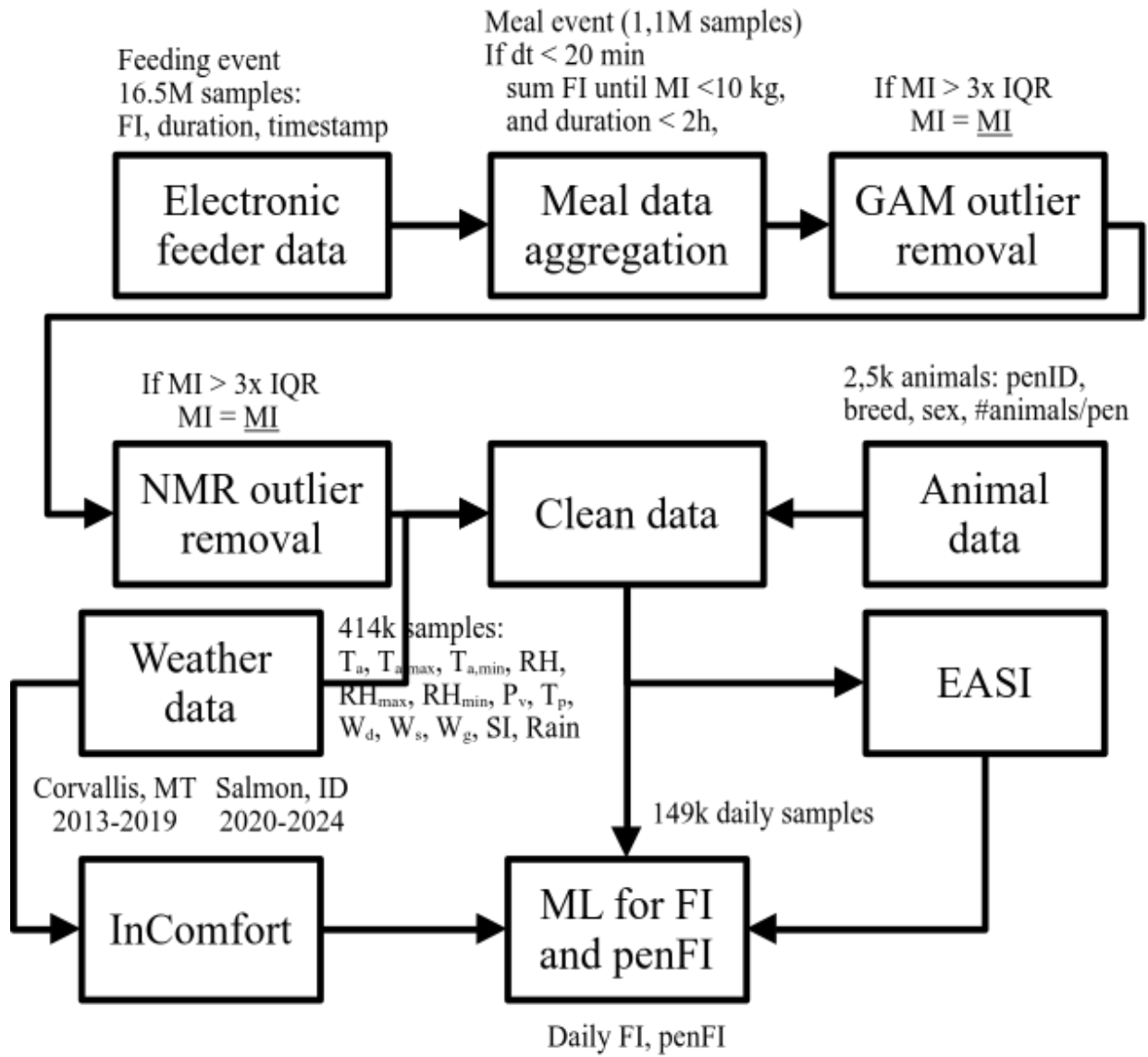

Figure 1. From the electronic feeder, we obtained 16.5M samples that contained feed intake (FI, kg), FI duration, and timestamp of starting FI. Individual IF data separated by less than 20 min (dt < 20 min) were aggregated into meal data (1,1M samples), considering a maximum meal duration of 2h and maximum meal intake (MI, kg) of 10 kg. First, outliers removal used Generalized Additive Models (GAM) and substituted samples that had MI deviation from predicted (MI) greater than 3 interquantile range (IQR). The same outlier removal procedure was repeated for Nonlinear Monotonic Regression (NMR) with spline transformation. Clean data included meal data and weather data (from Corvallis, MT, weather station from 2013-2019; from Salmon, ID, weather station from 2020-2024): air temperature ($T_a$, °C), daily maximum $T_a$ ($T_{a,max}$, °C), daily minimum $T_a$ ($T_{a,min}$, °C), relative humidity (RH, %), daily maximum RH ($RH_{max}$, %), daily minimum RH ($RH_{min}$, %), mean vapor pressure ($P_v$, kPa), mean dew point temperature ($T_p$, °C), mean wind direction ($W_d$, °), mean wind hourly speed ($W_s$, m/s), wind peak gust ($W_g$, m/s), solar irradation (SI, W/m$^2$), and precipitation (Rain, mm). Using weather data, InComfort index was developed. Using clean data, EASI index was developed. Finally, daily FI and pen FI (penFI, kg; 149k samples) were predicted using machine learning algorithms (ML).

## 2.1. Experimental data

Data came from 19 feed efficiency/feed intake trials (2,546 animals; 16,568,851; 2013-2024; Fall, Winter, and Spring seasons) conducted at Nancy M. Cummings Research Extension & Education Center (Carmen, ID, USA) feedlot facility (4 pens of ±970 m$^2$ each; up to 50 yearling cattle per pen), equipped with GrowSafe systems to measure individual animal feed intake.

Animal meal duration (capped at 2 hours and/or 10 kg) was defined as the sum of feeding events separated by less than 20 minutes to account for misreadings due to the animal moving its head away from the sensor (1,189,756). Outliers were initially detected as deviations exceeding 3x the interquartile range from a generalized additive model estimation (dependent variables: meal duration, meal events per day, initial and final time of meal events) and replaced with model estimates. A second outlier detection and replacement step was applied using nonlinear monotonic regression with spline transformation following the same GAM procedure.

## 2.2. Meteorological data

Meteorological data (15-min resolution; 414,912 samples) were obtained from AgriMet Network (Palmer, 2011): Salmon, Idaho Station (SLMI – Latitude: 45.277772° N; Longitude: 113.883958° W; Elevation: 1164.74 m; 2020-2024) and Corvallis, Montana Station (COVM – Latitude: 46.33333° N; Longitude: 114.08333° W; Elevation: 334.16 m; 2013-2019; ~160 km distant form Salmon station). The following environmental variables were selected: 1) air temperature, 2) daily maximum air temperature, 3) daily minimum air temperature, 4) relative humidity, 5) daily maximum relative humidity, 6) daily minimum relative humidity, 7) daily mean vapor pressure, 8) daily mean dew point temperature, 9) daily mean wind direction, 10) daily mean wind speed, 11) daily wind peak gust, 12) solar radiation, and 13) precipitation. Time-lagged variables (1 to 6 days; 8,736 new variables) were generated. To prevent modeling bias, all variables were normalized to ensure homogeneity of scale.

## 2.3. Environmental indices

To synthesize the effects of meteorological variables on feed intake, two indices were developed: InComfort and EASI. InComfort-Index was calculated solely from meteorological variables to quantify thermal comfort. In contrast, EASI-Index was developed as a hybrid index by integrating both meteorological variables and the response variable (feed intake), enabling the index to directly reflect animal performance under varying conditions. Principal component analysis was used to reduce dimensionality (unsupervised learning) and eliminate redundancy and multicollinearity of meteorological variables while preserving 95% data variance (110 principal components). Indices were scaled to a [-1, 1] range, where values near -1 indicate cold stress, values near 0 indicate the thermal comfort zone, and values near +1 indicate heat stress.

Indices were segmented into 5 clusters using k-means clustering (Pedregosa et al., 2011; silhouette scores: InComfort-Index = 0.5636; EASI-Index = 0.5398) to define interpretable thermal comfort classes: cold stress, slight cold stress, comfortable, slight heat stress, and heat stress. InComfort-Index was computed using principal components scaled from 0 to 1 and linearly normalized using Membership Function Value Analysis. Normalized components were weighted by their eigenvalues and summed to obtain the index. For EASI-Index, gradient boosted decision trees were used to predict feed intake from principal components. Training was performed on 50% of the dataset (stratified random sampling), and hyperparameters were selected to balance bias and variance. Performance was evaluated using coefficient of determination ($R^2$) and Root Mean Squared Error (RMSE).

## 2.4. Machine learning feed intake prediction

To predict individual animal daily intake (149,278 samples), four machine learning algorithms were evaluated: Extreme Gradient Boosting (XGBoost, Chen et al., 2015), Categorical Boosting (CatBoost, Prokhorenkova et al., 2018), Light Gradient Boosting Machine (LightGBM, Ke et al., 2017), and Bayesian Ridge Regression. Predictors included animal characteristics (e.g., breed, sex, number of animals), feeding behavior traits (e.g., number of meals, meal duration, feeding events), meteorological variables, InComfort-Index, and EASI-Index. Pen-level feed intake was estimated using the best performing model, including the best individual animal feed intake prediction as an explanatory variable. To maximize R² and minimize RMSE, training was performed on 50% of the data, and hyperparameters were optimized using 3-fold cross-validation. SHapley Additive exPlanations (SHAP) analysis was employed to determine relative importance of predictors and to explore model interpretability from the best performing model (Lundberg & Lee, 2017).

# 3. RESULTS AND DISCUSSION

## 3.1. Experimental data

As shown in Fig. 2, outliers were effectively identified and replaced with predicted values. Using the clean dataset (Fig. 3), feeding behavior traits and feed intake were reliably quantified. On average, animals spent 2h:12min ± 48 min eating per day, with meal intervals of 2h:39min ± 51min. This corresponded to approximately 8 meals daily, with a mean feed intake of 11.89 ± 3.08 kg/day. These eating times are consistent with previous reports for dairy (DeVries et al., 2003) and beef cattle (DelCurto-Wyffels et al., 2021). Across the 2,546 animals analyzed, the large standard deviations highlight pronounced individual animal variability, even under the same conditions. This variability likely reflects differences in individual eating patterns (Løvendahl and Munksgaard, 2016) and/or variation in the ability to cope with thermal stress (Montanholi et al., 2010), rather than being a data artifact.

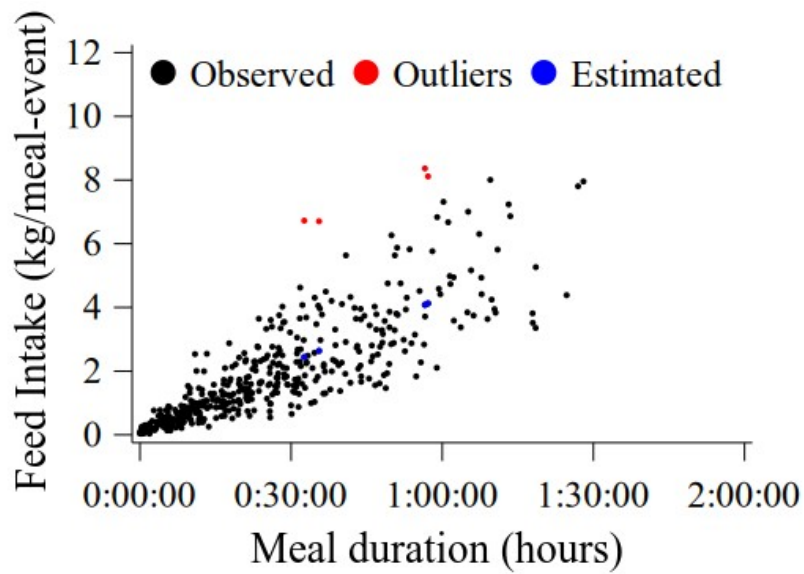

Figure 2. Example of outlier detection for one experimental animal. Feed intake per meal (kg/meal-event) is plotted against meal duration (hh:mm:ss). Red circles indicate detected outliers, and blue circles show values replaced with predicted values.

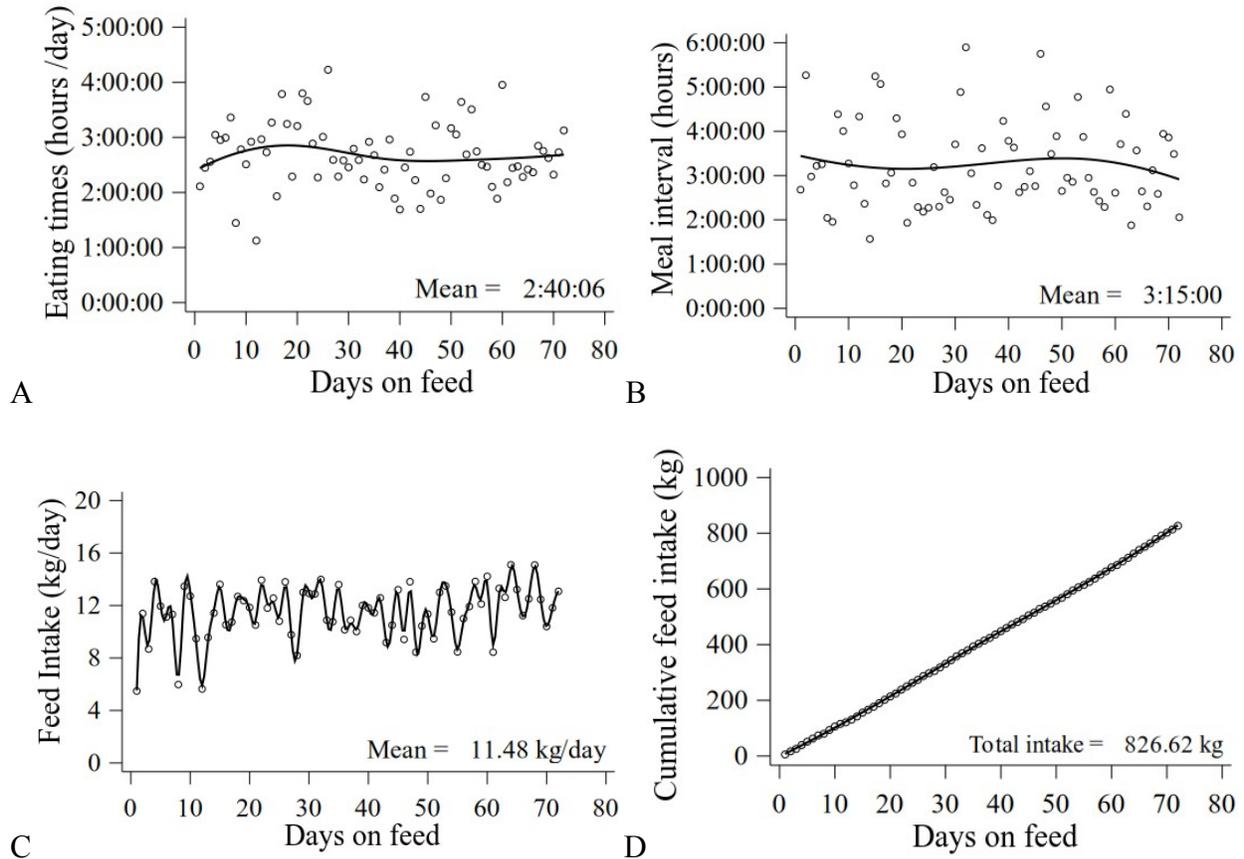

Figure 3. Feeding behavior traits and feed intake for one experimental animal. (A) Total eating time (hours/day), (B) meal interval (hh:mm:ss), (C) feed intake (kg/day), and (D) cumulative feed intake (kg) plotted against days on feed (days). Solid lines represent predicted values using generalized additive models.

## 3.2. Meteorological data

Meteorological data from the Salmon weather station (2020–2024) were complemented with Corvallis weather station data (2013–2019; ~160 km distant from Salmon station). As shown in Figs. 4 and 5, both sources exhibited similar temporal patterns, supporting their combined use.

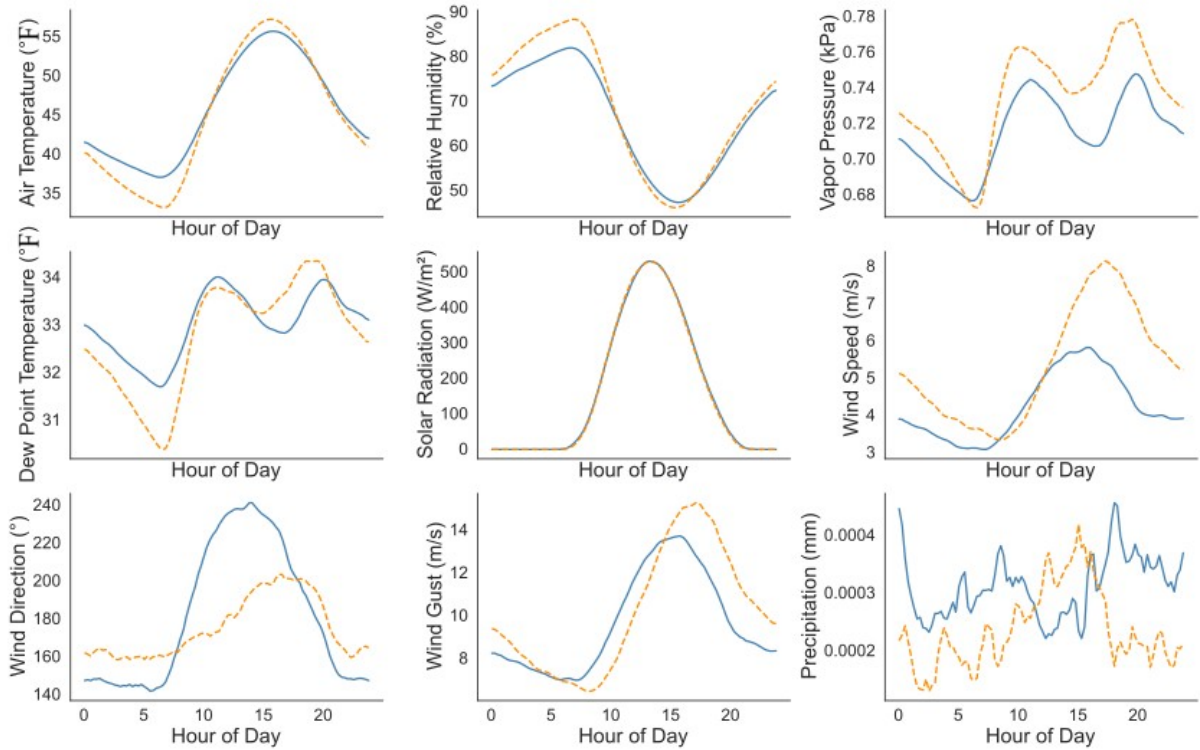

Figure 4. Diurnal variation of meteorological variables recorded by Corvallis (continuous blue lines; 2013–2019) and Salmon (dashed orange lines; 2020–2024) weather stations. Means were smoothed using a centered 3-hour moving average.

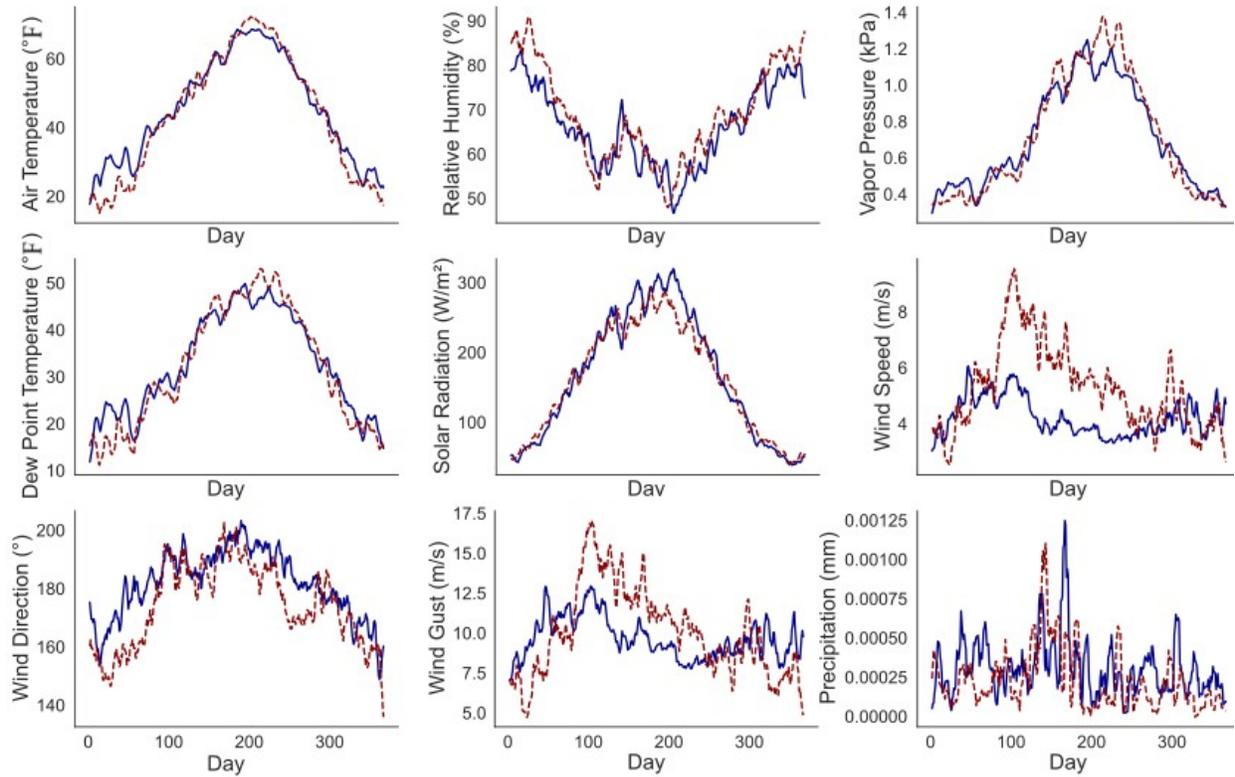

Figure 5. Yearly variation of meteorological variables recorded by Corvallis (continuous blue lines; 2013–2019) and Salmon (dashed maroon lines; 2020–2024) weather stations. Means were smoothed using a centered 7-days moving average.

Meteorological data indicated that animals were exposed to a continental climate with a wide thermal variation. Air temperature showed significant seasonal variability, ranging from -30.3 °C to 42.5 °C (mean = 7.1 °C; amplitude 54.5 °C). Relative humidity averaged 68.4%, with seasonal variability exceeding 80% throughout the year. Vapor pressure (mean = 0.73 kPa) followed daily and annual thermal variations, with higher values during warmer hours and days. Dew point temperature mirrored the dynamics of air temperature and humidity (mean = 0.4 °C). Solar radiation peaked around midday and during summer, averaging of 165.0 W/m² and reaching a maximum of 1,217.4 W/m². Wind speed averaged 5.34 m/s, predominantly from 175.3°, showing substantial angular dispersion throughout the year.

No strong correlations were found between feed intake and individual meteorological variables (<0.072). This contrasts with previously reported correlations up to 0.31 (Mujibi et al., 2010). These findings suggest that individual meteorological variables show inconsistent associations with feeding behavior and feed intake throughout the year, underscoring for integrated environmental indices that capture cumulative thermal stress and reliable predict feed intake patterns under variable and sometimes extreme conditions.

**3.3. Environmental indices**

Fig. 6 shows the seasonal patterns of the two environmental indices throughout the year. InComfort-Index classified thermal comfort categories in agreement with expected seasonal variation (cold stress in winter, heat stress in summer), indicating good prediction capability for assessing daily thermal comfort. In contrast, EASI-Index showed less seasonal pattern, with thermal comfort categories distributed more evenly throughout the year. Because EASI-Index incorporates animal responses (feed intake from preceding days), this indicates that not all climatic variability directly translates into changes in feeding behavior or feed intake.

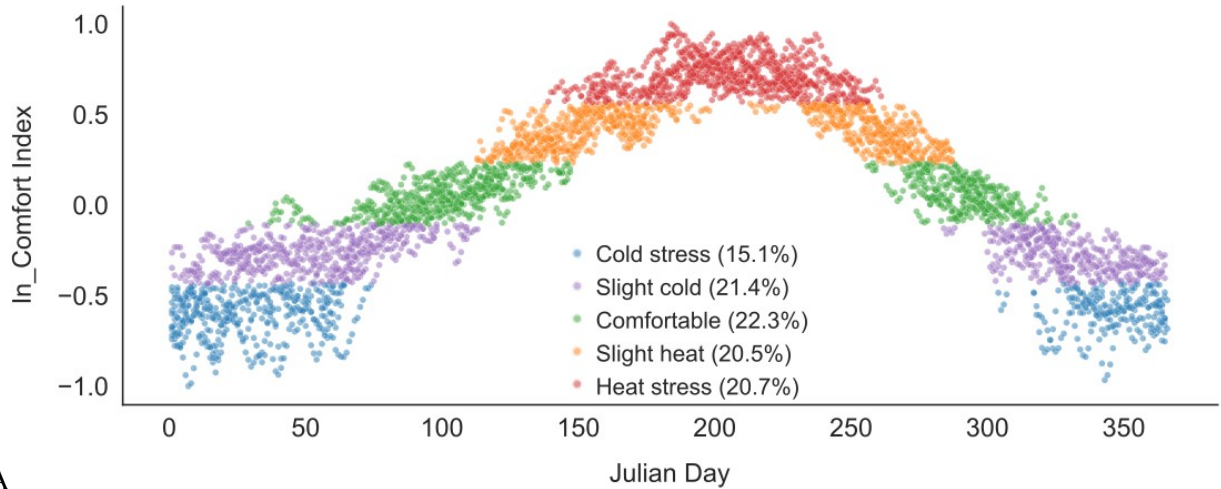

A

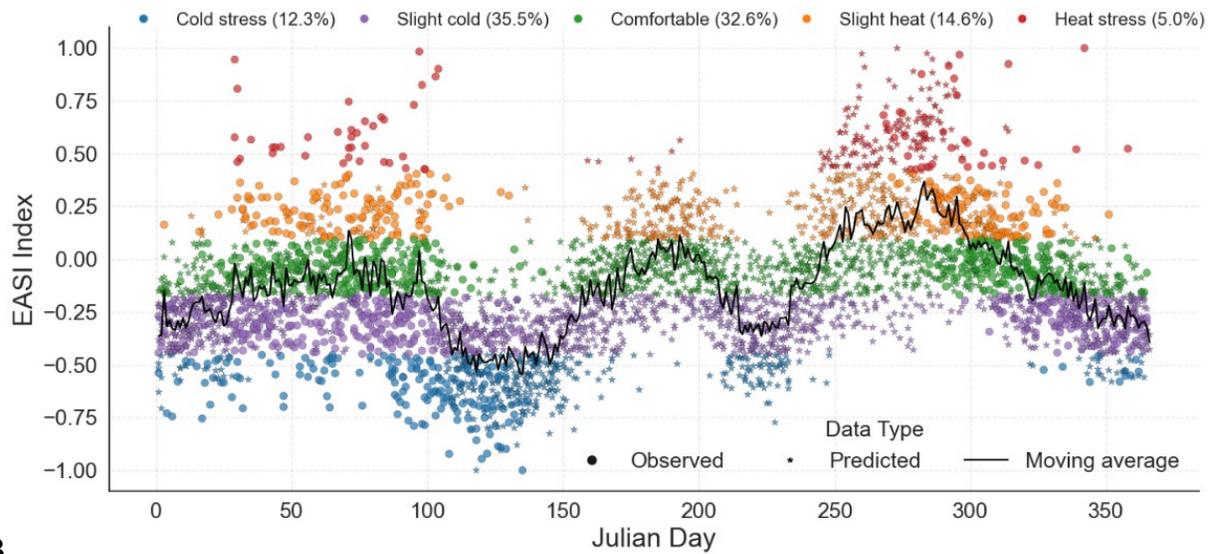

B

Figure 6. Environmental indices throughout the year. (A) InComfort-Index primarily reflects thermal conditions. (B) EASI-Index integrates thermal condition that affect feed intake responses. Thermal comfort categories classified according to each index: cold stress, slight cold stress, comfortable, slight heat stress, and heat stress.

InComfort-Index (Fig. 6A) was strongly correlated with air temperature (0.85), maximum temperature (0.80), minimum temperature (0.76), dew point temperature (0.71), and vapor pressure (0.68). It showed moderate correlations with relative humidity (-0.46), minimum relative humidity (-0.48), and solar radiation (0.42). Although these correlations indicate that InComfort-Index effectively reflects thermal environment conditions, the index showed no correlation with feed intake (0.005). This likely reflects its design, which emphasizes environmental conditions indicative of thermal comfort rather than cumulative physiological responses to environmental stress. These findings underscore the importance of developing environmental indices that capture cumulative effects of environmental conditions on animal responses such as feed intake.

By contrast, EASI-Index (Fig. 6B) was moderately correlated with feed intake (-0.580) but showed weak correlation with individual environmental variables (<0.13). This highlights the complexity, interdependence, and time-dependence of environmental stress effects on animal responses. For instance, the EASI-Index classified nearly half of the days (47.8%) as cold-stress, suggesting prolonged cold thermal challenge, which can increase feed intake, without necessarily leading to body weight gain.

### 3.4. Machine learning for feed intake prediction

Fig. 7 shows feed intake predictions from machine learning models. Among tested algorithms, XGBoost achieved the best performance with the lowest RMSE (training: 1.07 kg/day; validation: 1.38 kg/day) and highest $R^2$ (training: 0.879; validation: 0.798). CatBoost and LightGBM performed comparably, while Bayesian ridge regression performed less well. Unlike previous reports that noted limited predictive accuracy (ArunKumar et al., 2025), our approach maintained high prediction accuracy in the validation dataset, suggesting both robustness and lack of overfitting to the training data.

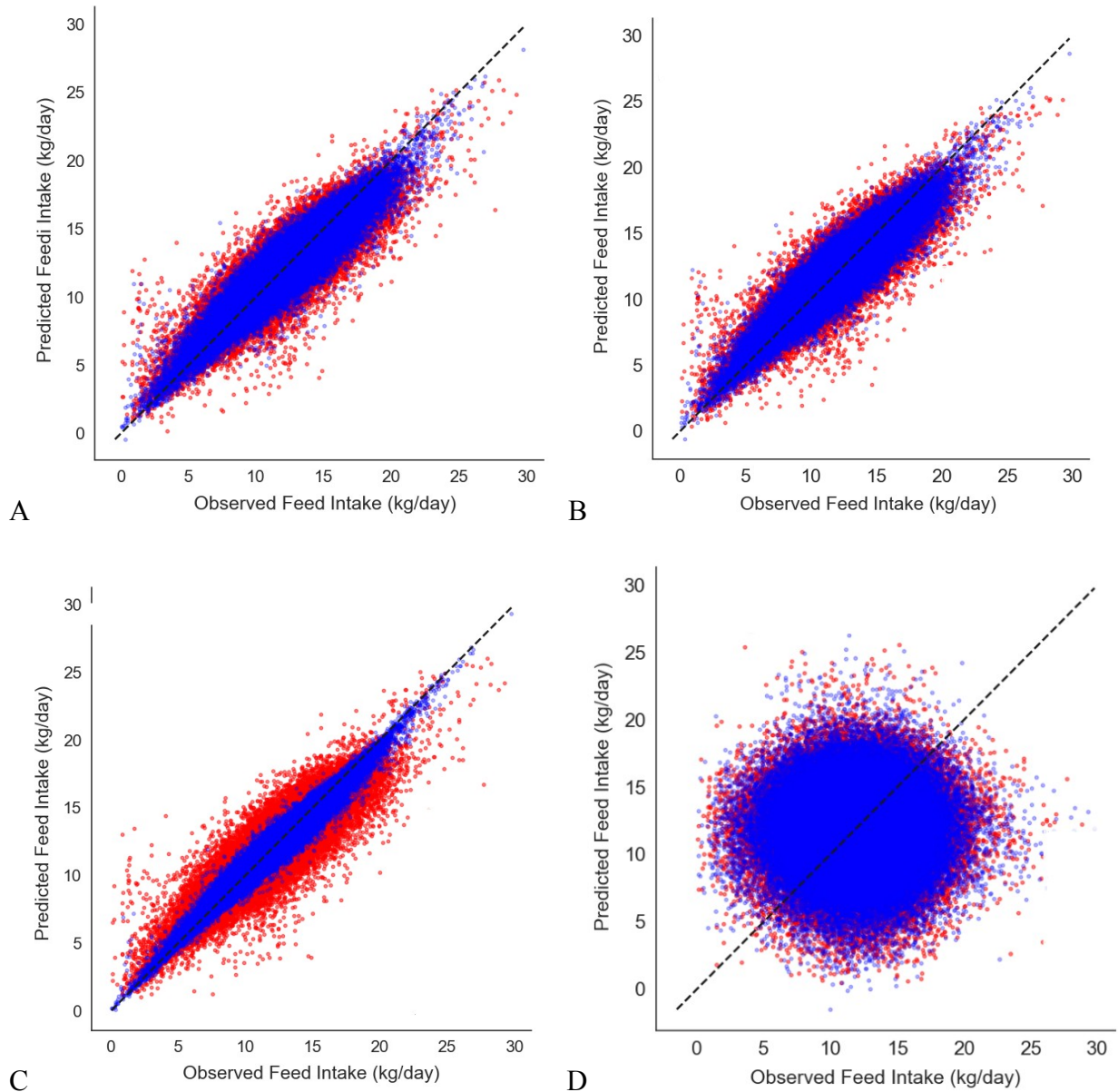

Figure 7. Performance of machine learning models to predict individual animal feed intake. (A) XGBoost (training RMSE: 1.0738; validation RMSE: 1.3852; training R2: 0.8789, validation

R2: 0.7978); (B) CatBoost (training RMSE: 1.1481; validation RMSE: 1.3913; training R2: 0.8615, validation R2: 0.7960); (C) LightGBM (training RMSE: 0.5740; validation RMSE: 1.3931; training R2: 0.9654, validation R2: 0.7955); (D) Bayesian ridge regression (training RMSE: 1.4350; validation RMSE: 1.4620; training R2: 0.6840, validation R2: 0.7740). Blue points represent training data; red points represent validation data; dashed line indicates the 1:1 line.

Relative importance analysis (Fig. 8) showed that the top 4 predictors together accounted for nearly 42% of total importance, indicating their strong influence on feed intake. The top 10 predictors accounted for nearly 70% of total relative importance, whereas several variables showed only marginal importance (<0.3%). Interestingly, the EASI-Index alone accounted for 7.4%, which was greater than the combined contribution of all individual environmental variables (<4%). This indicates the value of integrated indices over isolated environmental measures when predicting animal responses.

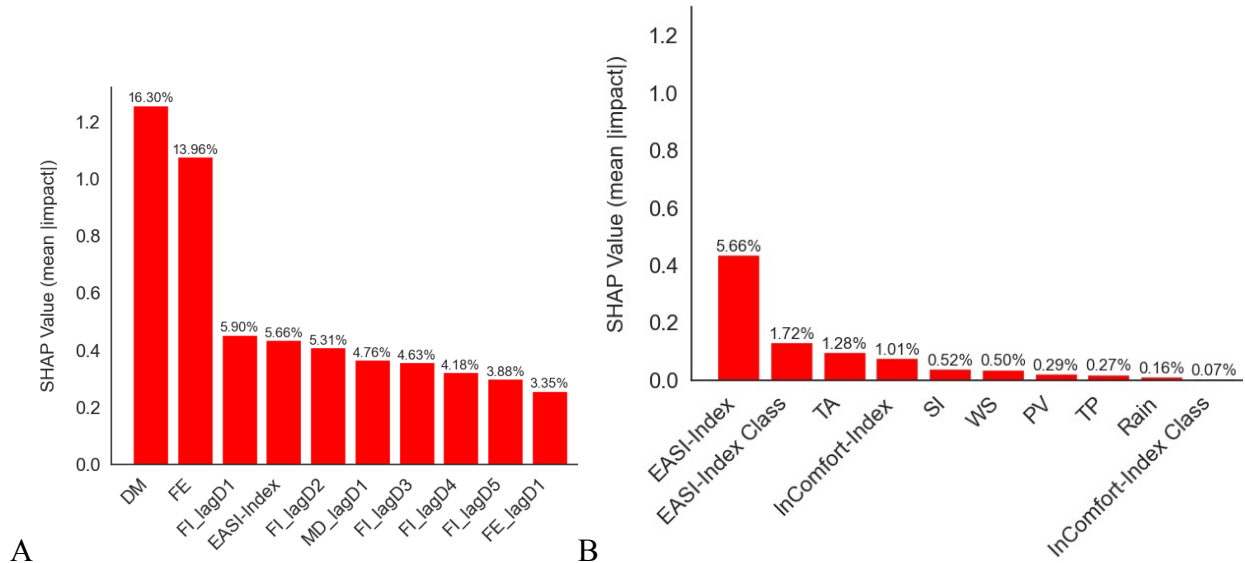

Figure 8. Relative importance of predictors in the best-performing model (XGBoost) calculated using SHAP analysis. (A) Top 10 predictors across all variables. (B) Top 10 environmental predictors. DM = meal duration; FE = feeding event frequency; FI_lagDx = feed intake of x previous days; MD_lagDx = meal duration of x previous days; TA = air temperature; SI = solar radiation; WS = wind speed; PV = vapor pressure; TP = dew point temperature.

SHAP analysis (Fig. 9) showed that longer meal duration, higher feeding event frequency, and greater prior-day feed intake positively influenced predictions, while higher EASI-index values negatively influenced predictions. These associations align with expected feeding behavior and physiology, as more frequent and longer meals increase feed intake, cumulative heat stress reduces feed intake, and behavioral patterns exhibit temporal carryovers from previous days (Collier and Gebremedhin, 2015). Our findings indicate the importance of considering integrated indices and incorporating prior-day observations when predicting feed intake.

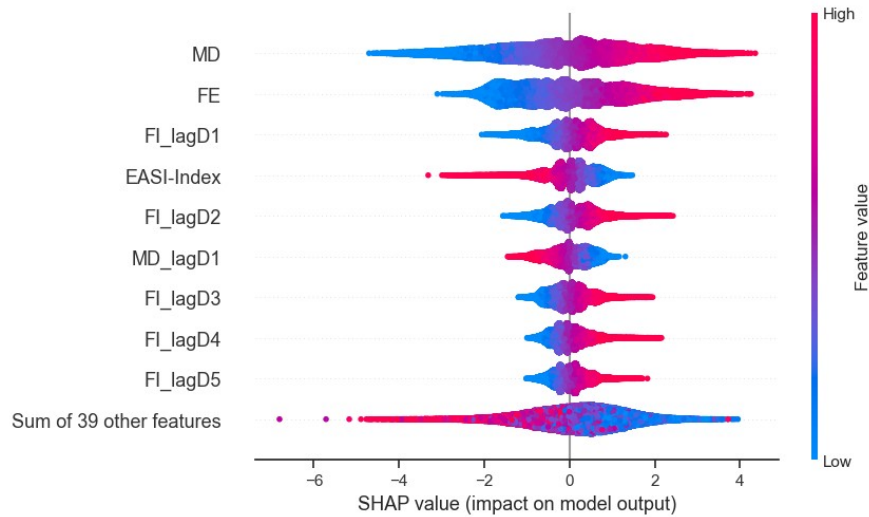

Figure 9. SHAP summary plot for the best performing model (XGBoost) ranked by predictor relative importance (Fig. 8). Red points indicate relatively high predictor values; blue points indicate low values. Positive SHAP values indicate influence on increasing the predicted feed intake, whereas negative SHAP values indicate influence on decreasing the predicted feed intake. MD = meal duration; FE = feeding event frequency; FI_lagDx = feed intake of x previous days; MD_lagDx = meal duration of x previous days.

Fig. 10 shows feed intake prediction as a function of environmental indices. Simulated predictions revealed a nonlinear decline in feed intake with increasing EASI-Index values, reflecting the suppressive effects of heat stress. Interestingly, cold stress did not markedly reduce intake, likely because greater consumption compensates for thermoregulatory energy demands, but at the cost of reduced efficiency. By contrast, InComfort-Index showed little direct influence on feed intake. These results reinforce the value of indices that integrate cumulative animal responses rather than instantaneous meteorological conditions, given the complex and time-dependent nature of thermal stress. For EASI-Index, this means that stressful environments reflect not only immediate thermal challenges but also cumulative exposures that constrain feed intake.

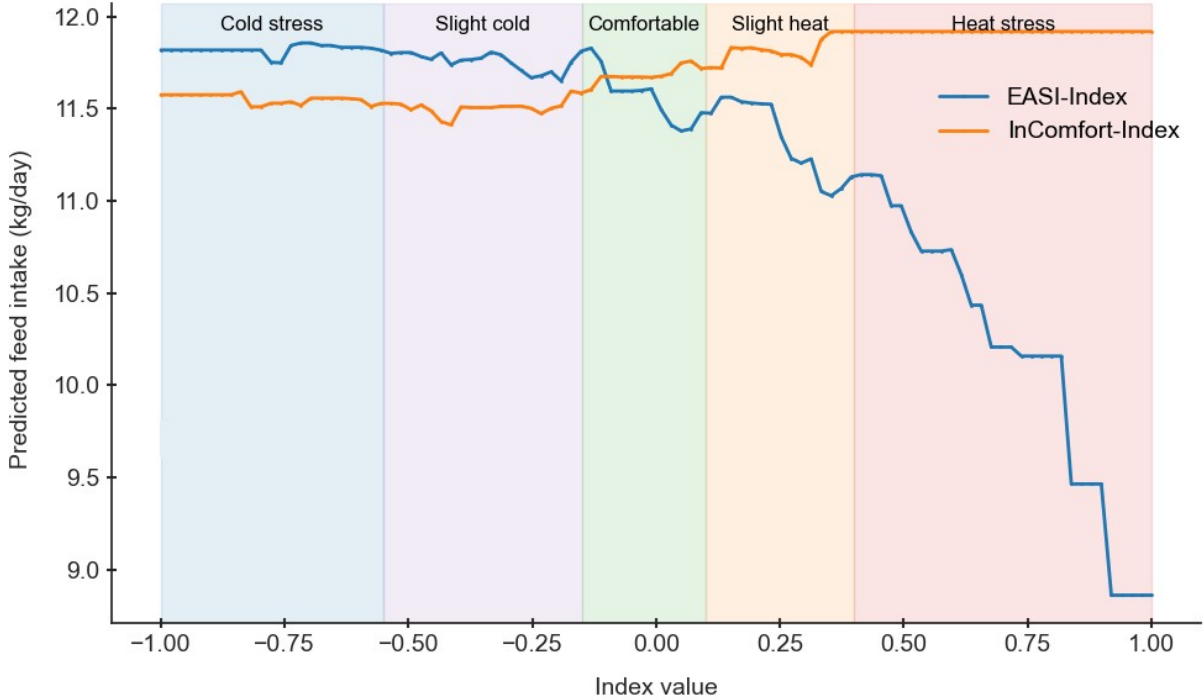

Figure 10. Simulation of predicted feed intake as a function of environmental indices using the best-performing model (XGBoost), with other predictors set to their mean values. EASI-Index is shown as a continuous blue line and InComfort-Index as a continuous orange line. Thermal comfort categories were established according to EASI-Index classification.

Finally, pen-level feed intake predictions outperformed individual animal-level predictions (Fig. 11). The XGBoost model predicted pen-level feed intake with an error of only 140 g/(day-animal), consistent with the Law of Larger Numbers (Ross, 2014), where individual animal prediction errors average out at aggregate levels. This finding supports practical application of the methodology for pen-level management decisions in feedlot operations.

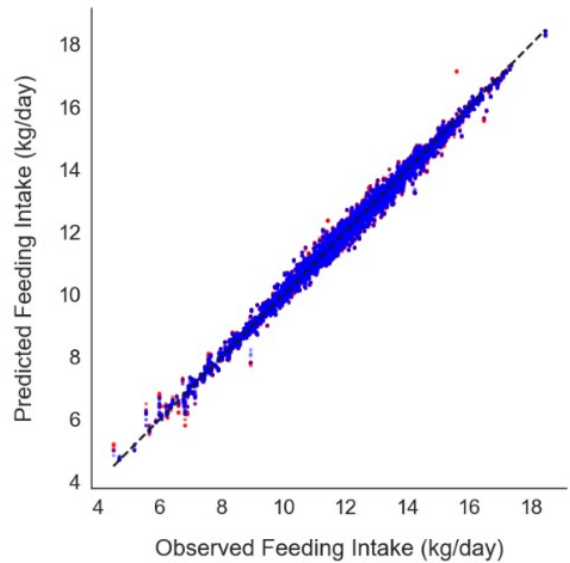

Figure 11. Performance of the best prediction model (XGBoost; training RMSE: 0.1545; validation RMSE: 0.1740; training R2: 0.9937, validation R2: 0.9920) for pen-level feed intake. Blue points represent training data; red points represent validation data; dashed line represents 1:1 line.

The proposed AI-based framework demonstrated sufficiently low errors (1.38 kg/day at animal-level and 0.14 kg/day-animal at pen-level) to enable several real-time precision management strategies, including automated feed delivery systems that precisely offer only the pen daily consumption, minimizing feed waste. Reduced feed waste directly translate into reduced production cost and, consequently, improved profit margins. In addition, evaluating the differences between predicted and consumed feed allow for early detection of health issues, which are intrinsically connected to lower feed intake. Furthermore, the environmental indices can be used to evaluate the cumulative effect of environmental stressors on feed intake, offering alerts in case of deviations or required interventions, such as turning on fans, sprinkles, or providing shades to the alleviate heat stress impact.

## 4. CONCLUSION

We developed and validated an AI-based framework to predict longitudinal feed intake in feedlot cattle that addresses limitations of traditional environmental indices. Through integrating Big Data Analytics (11 years of longitudinal data; >16.5M samples), advanced ML algorithms (XGBoost), and two novel environmental indices (InComfort-Index and EASI-Index), high accurate predictions were obtained: RMSE of 1.38 kg/day for individual animals and 0.14 kg/day-animal for pen-level aggregation. InComfort-Index classified thermal conditions based on thermal comfort but did not capture cumulative effects on animal responses, resulting in minimal influence on feed intake. In contrast, EASI-Index addressed this limitation by integrating cumulative animal responses and was strongly negatively correlated with feed intake, indicating that higher heat stress reduced feed intake, as expected. The proposed AI-based framework demonstrated strong predictive capacity for both individual animals and pen-level, indicating its suitability for deployment in commercial farms, advancing the field towards autonomous precision livestock farming systems that enhance both economic and environmental sustainability, through feed waste reduction, resource optimization, and climate-adaptive livestock management.

**Acknowledgments**

Funding: São Paulo Research Foundation (FAPESP; Proc. 2022/15473-0), Brazilian Coordination for the Improvement of Higher Educational Personal (CAPES; Proc. 88887.583358/2020-00, 88887.936679/2024-00), Brazilian National Council for Scientific and Technological Development (CNPq; Proc. 313341/2021-4, 446340/2024-3, 314878/2025-4, 442650/2025-6), and Mary Turner Endowment Awards.

Authors are thankful to the Nancy M. Cummings Research Extension & Education Center at the University of Idaho for providing access to the feedlot database.

**Authors contributions**

Conceptualization: ASCM & IAMAT; Data curation: JBH; Methodology: ASCM & HFMM; All authors: Writing, review, and editing.

**Data availability statement**

Research data is confidential

**Declaration of competing interests**

Nothing to declare.

**Declaration of generative AI and AI-assisted technologies in the writing process**

No generative AI or AI-assisted technologies were used during the preparation of this work.